\documentclass[letterpaper]{article}
\usepackage{aaai}
\usepackage{times}
\usepackage{helvet}
\usepackage{courier}
\usepackage{amsthm}
\usepackage{amsmath}
\usepackage{amssymb}
\usepackage{color}
\usepackage{graphicx}
\usepackage{dsfont}
\usepackage{algorithmic}
\usepackage{algorithm}

\newcommand{\todo}[2][]{}
\newcommand{\missingfigure}[2][]{}
\newtheorem{theorem}{Theorem}
\newtheorem{definition}{Definition}

\newtheorem{remark}{Remark}

\newcommand{\beqan}{\begin{eqnarray*}}
\newcommand{\eeqan}{\end{eqnarray*}}

\newcommand{\R}{{\mathbb R}}

\newcommand{\E}{\mathbb{E}}

\newcommand{\transpose}{^\mathsf{\scriptscriptstyle T}}

\DeclareMathOperator*{\argmin}{arg\,min}
\DeclareMathOperator*{\argmax}{arg\,max}

\newcommand{\cD}{\mathcal{D}}

\newcommand{\cG}{\mathcal{G}}

\newcommand{\cL}{\mathcal{L}}

\newcommand{\cN}{\mathcal{N}}

\newcommand{\cR}{\mathcal{R}}

\newcommand{\cV}{\mathcal{V}}
\newcommand{\cW}{\mathcal{W}}

\newcommand{\bq}{{\bf q}}

\newcommand{\bx}{{\bf x}}

\newcommand{\bI}{{\bf I}}
\newcommand{\bQ}{{\bf Q}}
\newcommand{\bX}{{\bf X}}

\newcommand{\bV}{{\bf V}}

\newcommand{\balpha}{{\boldsymbol \alpha}}

\newcommand{\bLambda}{{\boldsymbol \Lambda}}

\newcommand{\bff}{{\boldsymbol f}}

\newcommand{\specialcell}[2][c]{%
  \begin{tabular}[#1]{@{}c@{}}#2\end{tabular}}

\usepackage{natbib}

\begin{document}
%
\title{Spectral Bandits for Smooth Graph Functions \\with Applications in Recommender Systems}
\author{Tom\'a\v s Koc\'ak\\
SequeL team \\
Inria Lille \\
France\\
\And
Michal Valko\\
SequeL team \\
Inria Lille\\
France\\
\And
R\'emi Munos\\
SequeL team, Inria France\\
Microsoft Research\\ New England\\
\And
Branislav Kveton\\
Technicolor \\
Research Center\\
California\\
\And
Shipra Agrawal\\
Microsoft Research\\
Bangalore\\
India
}
\copyrightyear{2014}
\maketitle

\begin{abstract}
Smooth functions on graphs have wide applications in manifold and
semi-supervised learning. In this paper, we study a bandit problem where the
payoffs of arms are smooth on a graph. This framework is suitable for solving
online learning problems that involve graphs, such as content-based
recommendation. In this problem, each recommended item is a node and its
expected rating is similar to its neighbors. The goal is to recommend items that
have high expected ratings. We aim for the algorithms where the cumulative
regret would not scale poorly with the number of nodes. In particular, we
introduce the notion of an \emph{effective dimension}, which is small in
real-world graphs, and propose two algorithms for solving our problem that scale
linearly in this dimension. Our experiments on real-world
content recommendation problem show that a good estimator of user preferences
for thousands of items can be learned from just tens of node evaluations.
\end{abstract}

\section{Introduction}
\label{sec:intro}

\emph{A smooth graph function} is a function on a graph that returns similar
values on neighboring nodes. This concept arises frequently in manifold and
semi-supervised learning \citep{zhu2008semi-supervised}, and reflects the fact
that the solutions on the neighboring nodes tend to be similar. It is well-known
\citep{belkin2006manifold,belkin2004regularization} that a smooth graph function
can be expressed as a linear combination of the eigenvectors of the graph
Laplacian with smallest eigenvalues. Therefore, the problem of learning such as
a function can be cast as a regression problem on these eigenvectors. This is
the first work that brings this concept to bandits. In particular, we study a
bandit problem where the arms are the nodes of a graph and the expected payoff
for pulling an arm is a smooth function on this graph.

\begin{figure}[ht]
  \begin{center}
  \includegraphics[viewport = 112 239 497 558,clip,width=0.9\columnwidth]
{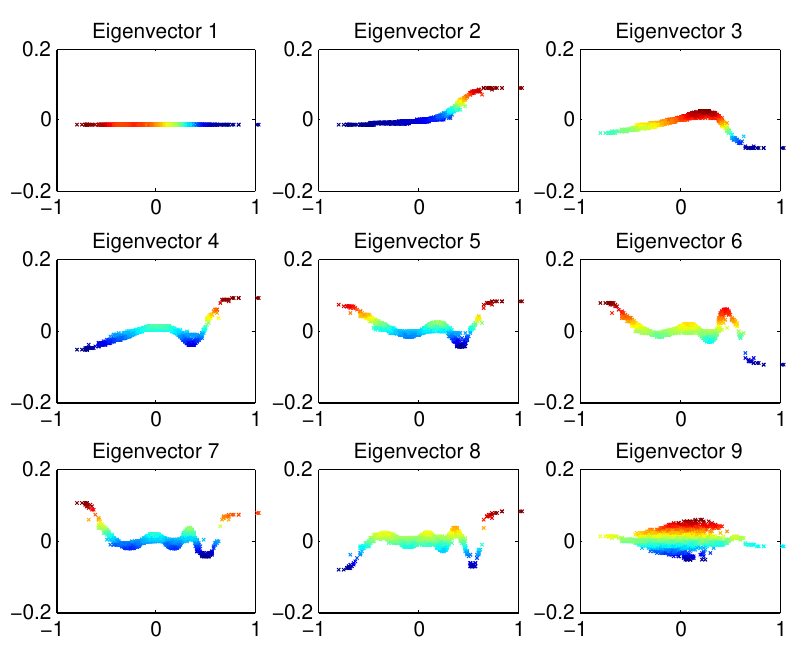}
 \vskip -0.5em
 \caption{Eigenvectors from the Flixster data corresponding to the smallest
few eigenvalues of the graph Laplacian projected onto the first principal
component of data. Colors indicate the values.}\vspace{-2em}
  \label{fig:flixster_eigenvectors}
  \end{center}
\end{figure}

Our work is motivated by a range of practical problems that involve graphs. One
potential application is \textit{targeted advertising} in social networks. In
this problem, the graph is a social network and our goal is to discover a part
of the network that is interested in a given product. Interests of people in a
social network tend change smoothly \citep{mcpherson2001birds}, because friends
tend to have similar preferences. Therefore, we can take advantage of this
structure and formulate this problem as learning a smooth preference function on
a graph.

Another application of our work are \textit{recommender systems}
\citep{jannach2010recommender}. In content-based recommendation
\citep{pazzani2007content}, the user is recommended items that are similar to the
items that the user rated highly in the past. The assumption is the user prefers
similar items similarly. The similarity of the items can be measured in many
ways, for instance by a nearest neighbor graph \citep{billsus2000learning}, where
each item is a node and its neighbors are most similar items. Therefore, the
problem of learning items that the user likes the most, can be naturally
formulated in our framework.

In this paper, we consider the following learning setting. The graph is known in
advance and its edges represent the similarity of the nodes in the graph. At
time $t$, we choose a node and then observe its payoff. In targeted
advertising, this may correspond to showing an ad and then observing whether
the person clicked on the ad. In content-based recommendation, this may
correspond to recommending an item and then observing the assigned rating. Based
on the payoff, we update our model of the world and then we proceed into
time $t + 1$. Since the number of nodes $N$ can be huge, we are interested in
the regime when $t < N$.

If the smooth graph function can be expressed as a linear combination of $k$
eigenvectors of the graph Laplacian, and $k$ is small and known, our learning
problem can be solved trivially using ordinary linear bandits
\citep{auer2002using,li2010contextual}. In practice, $k$ is problem specific and
unknown. Moreover, the number of features $k$ may approach the number of nodes
$N$. Therefore, a proper regularization is necessary so that the regret of the
learning algorithm does not scale with $N$. We are interested in the setting
where the regret is independent of $N$ and therefore our problem is non-trivial.

We make several major contributions. First, we formalize a bandit problem,
where the payoff of the arms is a smooth function on a graph. Second, we propose
two algorithms for solving this problem that achieve $d \sqrt{T\ln{T}}$ and $d\sqrt{
T\ln{N}}$ expected cumulative regret, where $d$ is the effective dimension of the
problem.  Finally, we evaluate both of the
algorithms on synthetic and real-world content-based recommendation problems.

\section{Setting}

Let $\cG$ be the given graph with the set of nodes $\cV$ and
denote $|\cV| = N$ the number of nodes.
Let $\cW$ be the $N \times N$ matrix of similarities $w_{ij}$ (edge
weights) and $\cD$ is the $N \times N$ diagonal matrix with the entries $d_{ii}
= \sum_j w_{ij}$.
The graph Laplacian of $\cG$ is defined as $\cL = \cD - \cW$.
Let $\{\lambda^\cL_k, \bq_k\}_{k=1}^N$ be the eigenvalues and eigenvectors
of $\cL$ ordered such that $ 0 = \lambda^\cL_1 \le \lambda^\cL_2 \le \dots
\le \lambda^\cL_N$.
Equivalently, let $\cL = \bQ \bLambda_\cL \bQ \transpose$ be the
eigendecomposition
of $\cL$, where $\bQ$ is an $N \times N$ \textit{orthogonal} matrix with
eigenvectors in columns.

In our setting we assume that the reward function  is a linear
combination of the eigenvectors. For any set of weights $\balpha$ let
$f_\balpha: \cV \to \R$ be the function
linear in the basis of the eigenvectors of $\cL$:
\[
 f_\balpha(v) = \sum_{k=1}^{N} \alpha_k (\bq_{k})_v = \langle \bx_v , \balpha
\rangle,
\]
where $\bx_v$ is the $v$-th row of $\bQ$, i.e., $(\bx_{v})_i = (\bq_{i})_v$.
If the weight coefficients of the true $\balpha^*$
are such that the large coefficients correspond
to the eigenvectors with the small eigenvalues
and vice versa, then $f_{\balpha^*}$ would be a smooth function on $\cG$
\citep{belkin2006manifold}.
Figure~\ref{fig:flixster_eigenvectors}
displays first few eigenvectors of the Laplacian
constructed from data we use in our experiments.
In the extreme case, the true $\balpha^*$ may be of the form
$[\alpha_1^*,\alpha_2^*, \dots, \alpha_k^*, 0, 0, 0]\transpose_N$
for some $k\ll N$. Had we known $k$ in such case, the
known linear bandits algorithm would work with the performance
scaling with $k$ instead of $D$. Unfortunately, first, we do not know $k$
and second, we do not want to assume such an extreme case (i.e.
$\alpha_i^* = 0$ for $i>k$). Therefore, we opt for the more plausible assumption
that the coefficients with the high indexes are small. Consequently, we deliver
algorithms with the performance that scale with the norm of $\balpha^*$
which expresses smoothness with respect to the graph.


%

The learning setting is the following. In each time step $t\le T$,
the recommender $\pi$ chooses a node $\pi(t)$ and obtains
a noisy reward $r_t =\bx_{\pi(t)} \transpose \balpha^* + \varepsilon_t$,
where the noise $\varepsilon_t$ is assumed to be $R$-sub-Gaussian for any $t$, i.e.,
$$\forall
\xi\in\R,\,\E[e^{\xi\varepsilon_t}]\leq
\exp\left(\frac{\xi^2R^2}{2}\right).$$
In our setting we have $\bx_v \in \R^D$ and $\|\bx_v\|_2\leq 1$
for all $\bx_v$.  The goal of the recommender is to minimize the cumulative
regret with respect to the strategy that always picks the best node
w.r.t.~$\balpha^*$.
Let $\pi(t)$ be the node picked (referred to as 
\textit{pulling an arm}) by an algorithm $\pi$ at time $t$.
The cumulative (pseudo) regret of $\pi$ is defined as:
\vspace{-.25em}
$$R_T  = T \max_v  f_{\balpha^*}(v) -  \sum_{t=1}^T f_{\balpha^*}(\pi(t)).$$
We call this bandit setting \textit{spectral}
since it is built on the spectral properties of a graph.
Compared to the linear and multi-arm bandits,
the number of arms
 $K$ is equal to the number of nodes $N$ and also to the
dimension of the basis $D$ (each eigenvector is of dimension $N$).
However, a regret that scales with $N$ or $D$
that can be obtained using those settings is not acceptable because the number
of nodes can be large.
While we are mostly interested in the setting with $K=N$,
our algorithms and analyses can be applied for any finite $K$.

\section{Effective dimension}
\label{sec:introeffd}

In order to present our algorithms and analyses
we introduce a notion of \textit{effective dimension} $d$. We keep using capital
$D$ to denote the ambient dimension, which is equal to $N$ in the spectral
bandits setting. Intuitively, the effective dimension is a proxy
for number of relevant dimensions. We first provide a
formal definition and then discuss its properties.

In general, we assume there exists a diagonal matrix $\bLambda$ with the
entries $0<\lambda=\lambda_1\leq
\lambda_2\leq \dots\leq \lambda_N$ and
a set of $K$ vectors $\bx_1,\dots, \bx_K\in\R^N$ such that $\|\bx_i\|_2\leq 1$
for all $i$. For the smooth graph functions, we have $K = N$.
Moreover since $\bQ$ is an orthonormal matrix, $\|\bx_i\|_2 = 1$.
Finally, since the first eigenvalues of a graph Laplacian is always zero,
 $\lambda^\cL_1 = 0$, we use $\bLambda = \bLambda_{\cL} + \lambda \bI$,
in order to have $\lambda_1 = \lambda$.

\begin{definition}\label{def:effectived}
 Let the \textbf{effective dimension} $d$ be the largest $d$ such
that:
$$ (d-1) \lambda_d  \leq \frac{T}{\log(1 + T /\lambda)}$$
\end{definition}

\begin{figure}[ht]
 \begin{center}
\includegraphics[width=0.44\columnwidth]{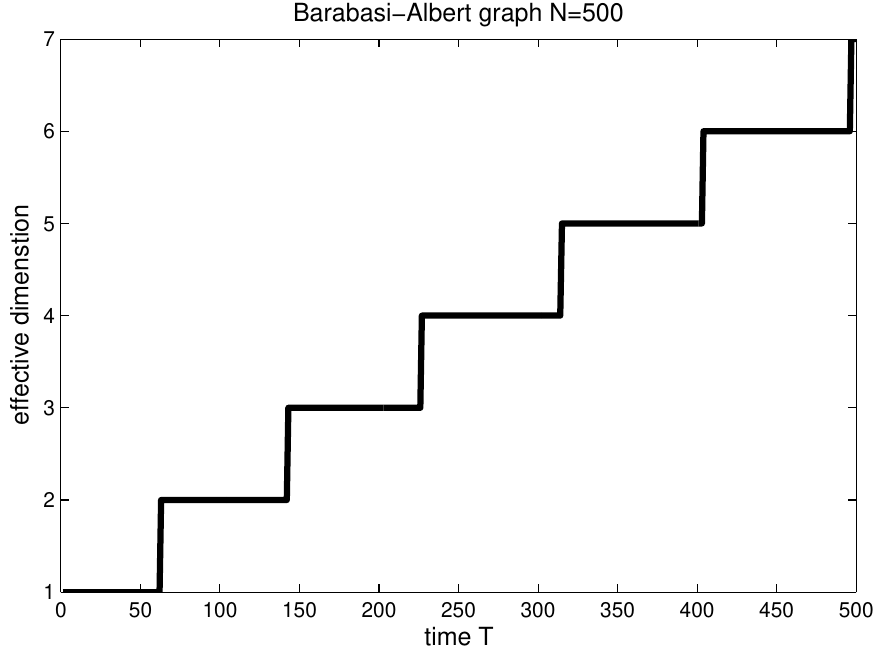}
\includegraphics[width=0.45\columnwidth]{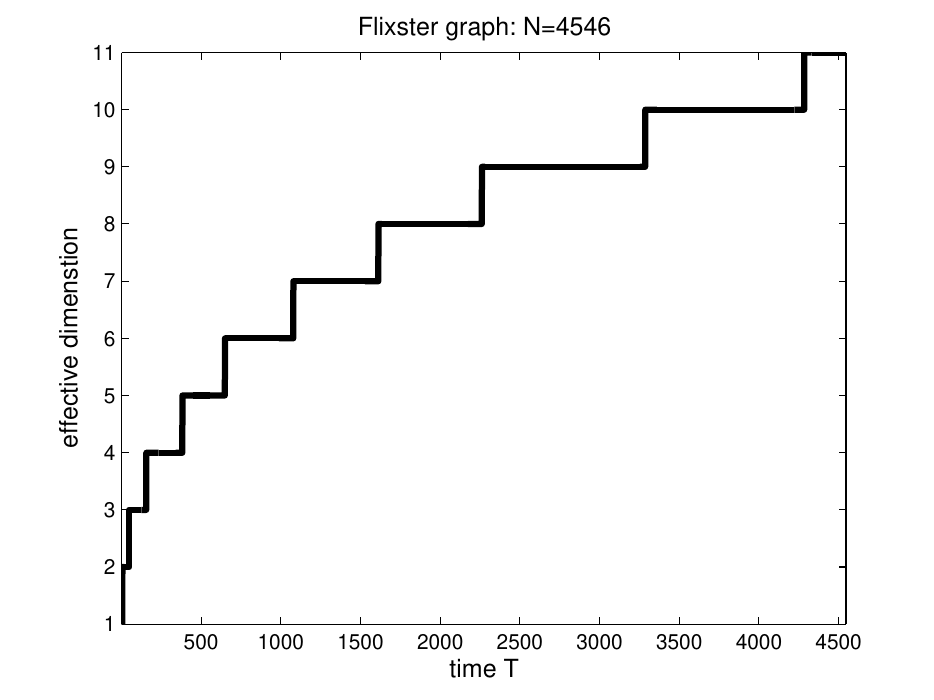}
 \vskip -0.5em
 \caption{Effective dimension $d$ in the regime $T < N$. The effective dimension
 for this data is much smaller than the ambient dimension $D$, which is 500
and 4546 respectively.}
 \label{fig:effd}
 \end{center}
\vspace{-0.5em}
 \end{figure}

The effective dimension $d$ is small when the coefficients $\lambda_i$ grow
rapidly above $T$. This is the case when the dimension of the space $D$ (and
$K$) are much larger than $T$, such as in graphs from social networks with very
large number of nodes $N$. In the contrary, when the coefficients are all small
then $d$ may be of order $T$ which would make the regret bounds useless.
Figure~\ref{fig:effd} shows how $d$ behaves compared to $D$ on the
generated and the real Flixster network graphs\footnote{We set $\bLambda$ to
$\bLambda_{\cL} + \lambda \bI$ with $\lambda = 0.01$,
where $\bLambda_{\cL}$ is the graph Laplacian of the respective graph.} that we
use for the experiments.

\section{Algorithms}
\label{sec:algo}
In this section we propose two algorithms. The first algorithm is based on 
LinUCB. For the second algorithm we use Thompson sampling (TS) approach.
\begin{algorithm}[t]
\caption{SpectralUCB}
 \label{alg:TUCB}
 \begin{algorithmic}
 \STATE {\bfseries Input:}
 \STATE \quad  $N:$ the number of vertices, $T:$ the number of pulls
\STATE \quad  $\{\bLambda_{\cL}, \bQ\}$ spectral basis of $\cL$
\STATE \quad  $\lambda, \delta:$ regularization and confidence parameters
\STATE \quad  $R,C:$ upper bounds on the noise and $\|\balpha^*\|_\bLambda$
\STATE {\bfseries Run:}
 \STATE  $\bLambda \gets \bLambda_{\cL} + \lambda \bI$
 \STATE  $d \gets \max\{d : (d - 1) \lambda_d \leq  T/\log(1 + T /\lambda) \} $
  \FOR{$t = 1$ {\bfseries to} $T$}
\STATE Update basis coefficients $\hat\balpha$:
\STATE \quad $\bX_t \gets [\bx_1,\dots,\bx_{t-1}]\transpose$
\STATE \quad $r \gets [r_1,\dots,r_{t-1}]\transpose$
\STATE \quad $\bV_t \gets \bX_t\bX_t\transpose + \bLambda$
\STATE \quad $\hat\balpha_t \gets \bV_t^{-1}\bX_t\transpose r$
\STATE $c_t \gets 2 R\sqrt{ d \log(1 + t/\lambda) + 2\log(1/\delta)} + C $
\STATE Choose node $v_{t}$ ($\bx_{v_{t}}$-th row of $\bQ$)
\STATE \quad $v_{t} \gets \argmax_v \left( f_{\hat\balpha}(v) + c_t \|\bx_v \|_{\bV_t^{-1}} \right)$
\STATE Observe reward $r_t$
\ENDFOR
 \end{algorithmic}
\end{algorithm}
\subsection{SpectralUCB}
The first algorithm we present is SpectralUCB (Algorithm~\ref{alg:TUCB})
which is based on LinUCB and uses the \textit{spectral
penalty}. 
For clarity, we set $\bx_{t} = \bx_{v_{t}} = \bx_{\pi(t)}$.
Here we consider regularized
least-squares estimate $\hat\balpha_t$ of the form:
$$
\hat\balpha_t = \argmin_{\balpha} \left( \sum_{\tau=1}^{t}\left[\langle \bx_\tau,
\balpha \rangle - r_\tau\right]^2 + \|\balpha\|_{\bLambda}  \right).
$$
A key part of the algorithm is to define the $c_t\|\bx\|_{\bV_t^{-1}}$
confidence widths for the prediction of the rewards.
We take advantage of our analysis 
to define $c_t$ based on the effective dimension $d$ which is specifically tailored to our setting.
The following theorem characterizes the performance
of SpectralUCB and bounds the regret as a function of
effective dimension~$d$.

\begin{theorem}[\cite{valko2014spectral}]
\label{thm:tucb}
Let $d$ be the effective dimension and $\lambda$ be the minimum
eigenvalue of $\bLambda$. If $\| \balpha^* \|_{\bLambda}\!\leq~\!C$
and for all $\bx_v$, $\langle \bx_v,\balpha^* \rangle \in [-1, 1]$,  then the
cumulative regret of SpectralUCB  is with probability at
least $1-\delta$ bounded as:
\begin{align*}
R_T &\leq  \left[8 R\sqrt{ d \log(1 + T/\lambda) + 2\log(1/\delta)} + 4C  + 4\right] \\
 & \quad  \times \sqrt{d T \log (1 + T/\lambda)}  
\end{align*}
\end{theorem}

\begin{remark}The constant $C$ needs to be such that $\| \balpha^*
\|_{\bLambda}\!\leq~\!\!C$.
If we set $C$ too small, the true $\balpha^*$ will lie outside of the region and far from $\hat\balpha_t$, causing the algorithm to underperform.
Alternatively $C$ can be time dependent, e.g.~$C_t = \log T$. In such case we
do not need to know an upperbound on $\| \balpha^* \|_{\bLambda}$
in advance, but our regret bound would only hold after some
$t$, where $C_t \geq \| \balpha^* \|_{\bLambda}$.
\end{remark}

\subsection{Spectral Thompson Sampling}

In this section, we use the Thompson Sampling approach to decide which arm to
play. Specifically, we will represent our
current knowledge about $\balpha^*$ as the normal distribution
$\cN(\hat\balpha(t),v^2\bV_t^{-1})$, where
$\hat\balpha(t)$ is our actual approximation of the unknown parameter $\balpha^*$ and
$v^2\bV_t^{-1}$ reflects our uncertainty about it for some constant $v$ specifically selected using our analysis (\cite{kocak2014spectral}).
As mentioned before, we assume that the reward function is a linear
combination of eigenvectors of $\cL$ with large coefficients corresponding to
the eigenvectors with small eigenvalues. We encode this assumption into our
initial confidence ellipsoid by setting $\bV_1 = \bLambda = \bLambda_\cL +
\lambda\bI$, where $\lambda$ is a regularization parameter.

After that, every time step $t$ we generate a sample
$\tilde\balpha(t)$ from distribution $\cN(\hat\balpha(t),v^2\bV_t^{-1})$
and chose an arm $\pi(t)$, that maximizes $\bx_i\transpose\tilde\balpha(t)$.
After receiving a reward, we update our estimate of $\balpha^*$ and the confidence of
it, i.e. we compute $\hat{\balpha}(t+1)$ and $\bV(t+1)$,
\begin{align*}
\bV_{t+1} &= \bV_t + \bx_{\pi(t)}\bx_{\pi(t)}\transpose	\\
\hat{\balpha}(t+1) &= \bV_{t+1}^{-1}\left(\sum_{i=1}^{t}\bx_{\pi(i)}r(i)\right).
\end{align*}

The computational advantage of SpectralTS in Algorithm~\ref{alg}, compared to
SpectralUCB, is that we do not need to compute the confidence bound for each
arm. Indeed, in SpectralTS we need to sample $\tilde\balpha$
which can be done in $N^2$ time (note that $\bV_t$
is only changing by a rank one update) and a maximum of
$\bx_i\transpose\tilde\balpha$ which can be also done in $N^2$ time.
On the other hand, in SpectralUCB, we need to compute a $\bV_t^{-1}$ norm
for each of $N$ context vectors which amounts to a $ND^2$ time.
Table~\ref{tab:comparison} (left) summarizes the computational
complexity of the two approaches. Finally note that in our setting
$D = N$, which comes to a  $N^2$ vs.~$N^3$ time per step.
We support this argument in Section~\ref{sec:experiments}. Finally
note that the eigendecomposition needs to be done only once in the beginning 
and since  $\cL$ is diagonally dominant,  this
can be done for $N$ in millions~\citep{koutis2010approaching}.

\begin{table}
\vspace*{-1.5em}
\begin{center}
\def\arraystretch{1.9}
 \begin{tabular}[r]{|r|cc|} \hline
 & \textit{Linear} & \textit{Spectral} \\ \hline
\specialcell{\textit{Optimistic Approach} \\[-1.2em] {\scriptsize $D^2N$  per
step
update}}
 & \specialcell{\textbf{LinUCB} \\[-1.2em]  {\scriptsize $D\sqrt{T\ln{T}}$}}
& \specialcell{\textbf{SpectralUCB} \\[-1.2em] {\scriptsize
$d\sqrt{T\ln{T}}$}} \\
 \hline
\specialcell{\textit{Thompson Sampling} \\[-1.2em] {\scriptsize $D^2+DN$ per
step
update}}
 & \specialcell{\textbf{LinearTS} \\[-1.2em]  {\scriptsize $D \sqrt{T \ln N}$}}
& \specialcell{\textbf{\textit{SpectralTS}} \\[-1.2em] {\scriptsize $d\sqrt{T
\ln N}$}} \\
 \hline
\end{tabular}
\vspace*{-0.5em}
\caption{Linear vs. Spectral Bandits}
\label{tab:comparison}
 \end{center}
\end{table}

\begin{algorithm}
   \caption{Spectral Thompson Sampling}
   \label{alg}
\begin{algorithmic}
   \STATE {\bfseries Input:}
   \STATE \quad $N$: number of arms, $T$: number of pulls
   \STATE \quad $\{\bLambda_\cL,\bQ\}$: spectral basis of graph Laplacian $\cL$
   \STATE \quad $\lambda$, $\delta$: regularization and confidence parameters
   \STATE \quad $R$, $C$: upper bounds on noise and $\|\balpha^*\|_\bLambda$
   \STATE {\bfseries Initialization:}
   \STATE \quad $v = R\sqrt{6d\ln((\lambda + T)/\delta\lambda)}+C$
   \STATE \quad  $\hat{\balpha} =  0_N$, $\bff=0_N$,
   $\bV = \bLambda_\cL + \lambda\bI_N$ 
   \STATE {\bfseries Run:}   
   \FOR{$t=1$ {\bfseries to} $T$}
   \STATE Sample $\tilde{\balpha} \sim \cN(\hat{\balpha},v^2\bV^{-1})$
   \STATE $\pi(t)\leftarrow \argmax_{a}\bx_a\transpose\tilde{\balpha}$
   \STATE Observe a noisy reward $r(t) = \bx_{\pi(t)}\transpose\balpha^* + \varepsilon$
   \STATE $\bff \gets \bff + \bx_{\pi(t)}r(t)$
   \STATE Update $\bV \gets \bV+ \bx_{\pi(t)}\bx_{\pi(t)}\transpose$
   \STATE Update $\hat{\balpha}\gets \bV^{-1}\bff $
   \ENDFOR
\end{algorithmic}
\end{algorithm}
We would like to stress that we consider the regime when $T < N$,
because we aim for applications with a large set of arms and we are
interested in a satisfactory performance  after just a few
iterations. For instance, when we aim to recommend $N$ movies, we
would like to have useful recommendations in the time $T < N$, i.e.,~before
the user saw all of them. The following theorem upperbounds the cumulative regret of
SpectralTS in terms of $d$.

\begin{theorem}[\cite{kocak2014spectral}]\label{mainTheorem}
Let $d$ be the effective dimension and $\lambda$ be the minimum eigenvalue
of $\bLambda$. If $\|\balpha^*\|_\bLambda\leq C$ and for all $\bx_i$,
$|\bx_i\transpose \balpha^*|\leq1$, then the cumulative regret of Spectral
Thompson Sampling is with probability at least $1-\delta$ bounded as
\begin{align*}
\cR_T\leq\,&\frac{11g}{p}\sqrt{\frac{4+4\lambda}{\lambda}dT\ln\frac{\lambda+T}{\lambda}}+\frac{1}{T}	\\
&+ \frac{g}{p}\left(\frac{11}{\sqrt{\lambda}}+2\right)\sqrt{2T\ln\frac{2}{\delta}},
\end{align*}
where $p = 1/(4e\sqrt{\pi})$ and
\begin{align*}
g =\, &\sqrt{4\ln TN}\left(R\sqrt{6d\ln\left(\frac{\lambda+T}{\delta\lambda}\right)}+C\right)	\\
&+R\sqrt{2d\ln\left(\frac{(\lambda+T)T^2}{\delta\lambda}\right)}+C.
\end{align*}
\end{theorem}

\begin{remark}
Substituting $g$ and $p$ we see that regret bound scales with
$d\sqrt{T\ln{N}}$. Note that $N=D$ could be exponential in $d$ and we need to
consider factor $\sqrt{\ln{N}}$ in our bound. On the other hand if $N$ is
exponential in $d$, then our algorithm scales with
$\ln{D}\sqrt{T\ln{D}}=\ln(D)^{3/2}\sqrt{T}$ which is even better.
\end{remark}

\section{Experiments}
\label{sec:experiments}
\begin{figure}[ht]
 \begin{center}
\includegraphics[width=0.49\columnwidth]{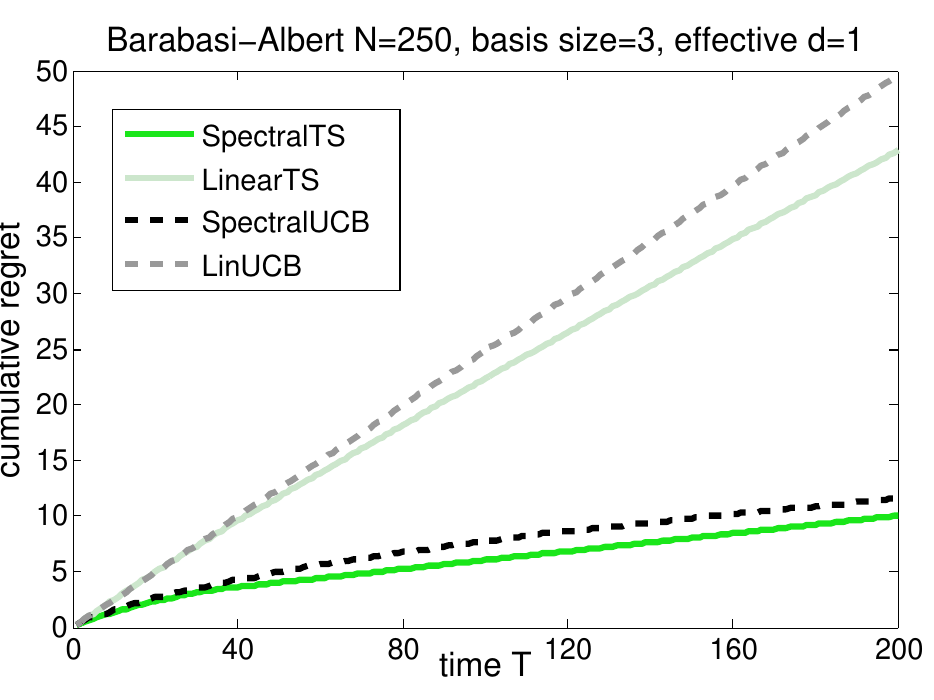}
\includegraphics[width=0.49\columnwidth] {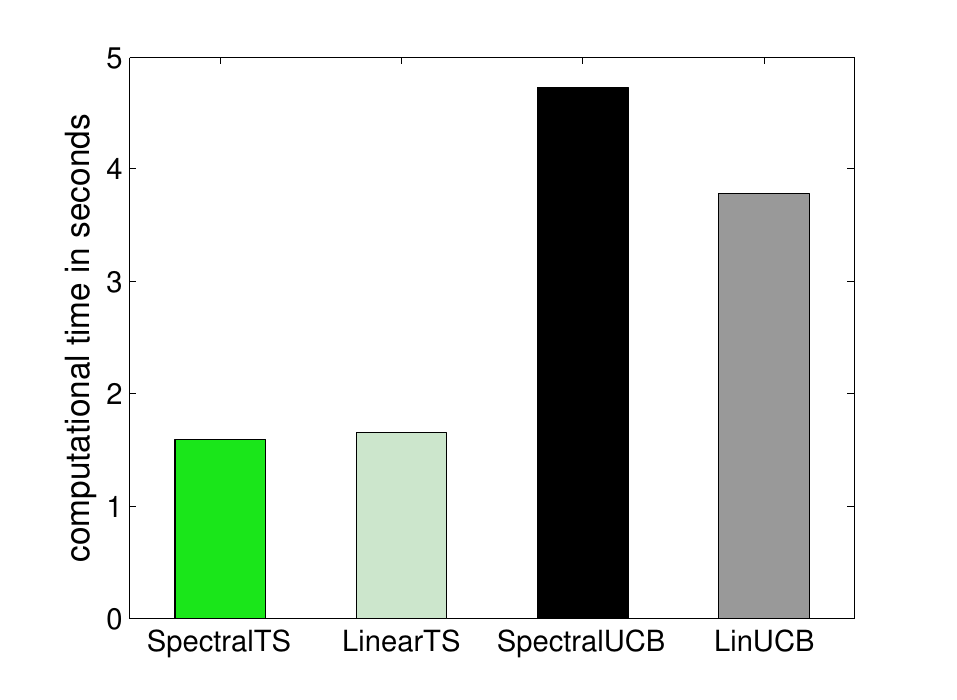}
\vspace{-1em}
\caption{Barab\'asi-Albert random graph results}
  \label{fig:syn}
\end{center}
\vspace{-0.25em}
 \end{figure}

The aim of this section is to give empirical
evidence that SpectralTS and SpectralUCB deliver better empirical
performance than LinearTS and LinUCB in $T<N$ regime.
For the synthetic experiment, we considered \textit{Barab\'asi-Albert} (BA)
model~(\citeyear{barabasi1999emergence}), known for its preferential
attachment property, common in real-world graphs.
We generated a random graph using BA model with $N = 250$ nodes and the degree
parameter $3$. For each run, we generated the weights of the edges uniformly
at random. Then we generated $\balpha^*$, a random vector of weights (unknown)
to algorithms in order to compute the payoffs and evaluated the cumulative
regret. The $\balpha^*$ in each simulation was a random linear combination
of the first 3 smoothest eigenvectors of the graph Laplacian.
In all experiments, we had $\delta = 0.001$, $\lambda=1$, and $R =
0.01$. We evaluated the algorithms in $T < N$ regime, where the linear bandit
algorithms are not expected to perform well. Figure~\ref{fig:syn} shows the
results averaged over 10 simulations. Notice that while the result of SpectralTS
are comparable to SpectralUCB, its computational time is much faster due the
reasons discussed
in Section~\ref{sec:algo}.
Recall that while both algorithms compute the least-square
problem of the same size, SpectralUCB has then to compute the confidence
interval for each arm.

\vspace{-.2cm}
\vspace{-0.4cm}

\begin{figure}[H]
\begin{center}
\includegraphics[width=.49\columnwidth]{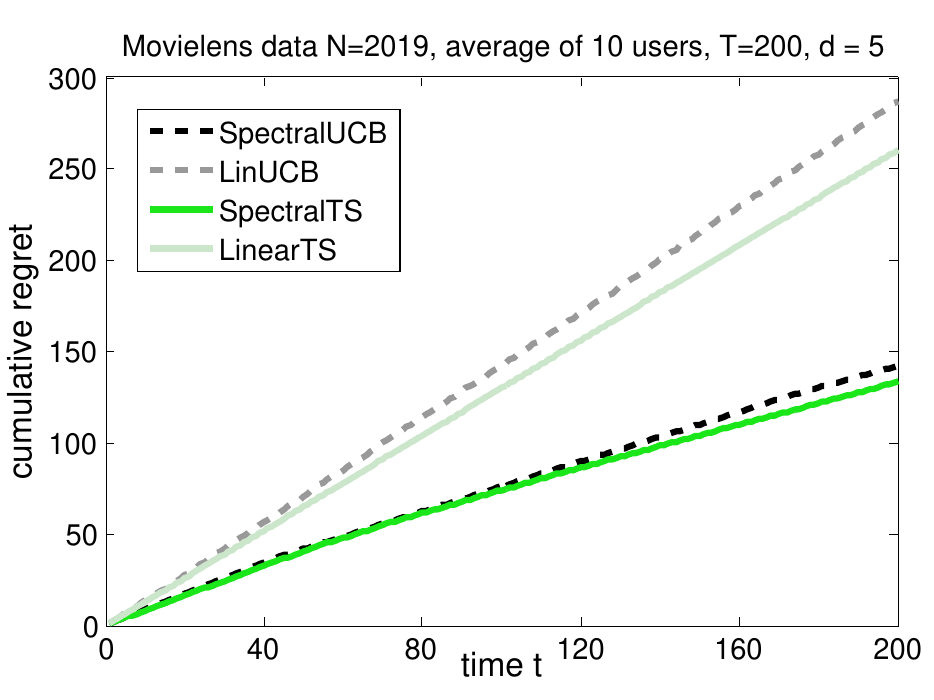}
 \includegraphics[width=0.49\columnwidth]{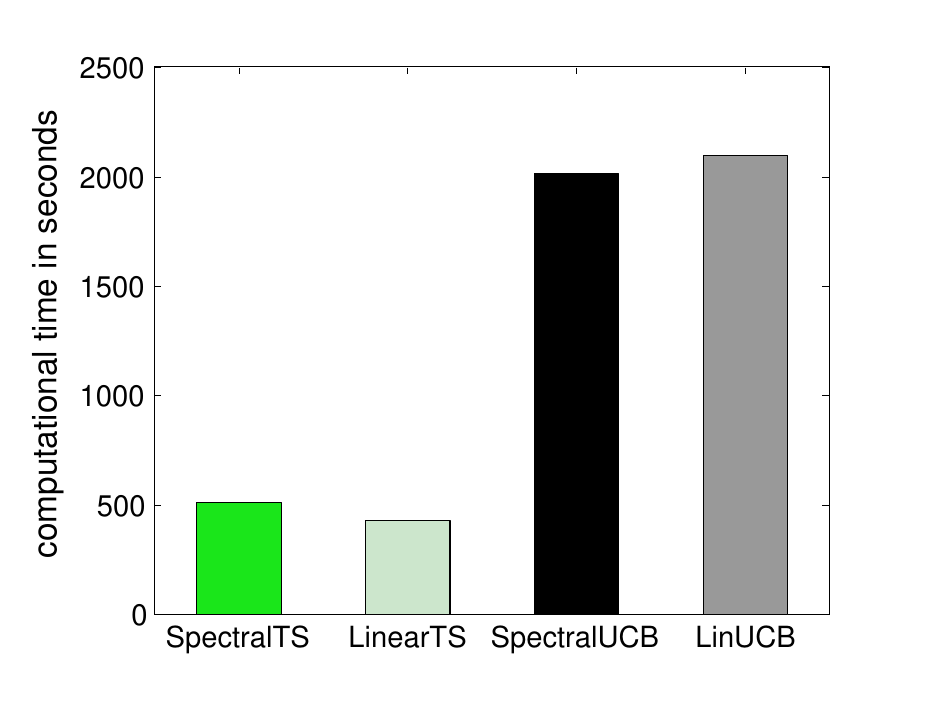}
\vspace{-1em}
\caption{MovieLens data results}
  \label{fig:movavg}
\end{center}
 \vspace{-0.25em}
 \end{figure}

Furthermore, we performed the comparison on the
MovieLens dataset \citep{movielens} of the movie ratings.
The graph in this dataset is the graph of 2019 movies edges corresponding
to the movie similarities. We completed the missing ratings by a low-rank matrix factorization
and used it construct a 10-NN similarity graph. For each user we have a graph function, unknown to the
algorithms, that assigns to each node, the rating of the particular user.
A detailed description on the preprocessing is deferred
to~\citep{valko2014spectral} due to limited space.
Our goal is then to recommend the most highly rated content.
Figure~\ref{fig:movavg} shows the MovieLens data results averaged
over 10 randomly sampled users. Notice that the results
follow the same trends as  for synthetic data.
Overall, both our spectral algorithms outperform their linear 
counterparts in terms of cumulative regret.

%

\vspace{-.2cm}



\newpage
\bibliography{library}
\bibliographystyle{plain}

\end{document}